\documentclass[runningheads]{llncs}

\usepackage{eccv}

\usepackage{eccvabbrv}

\usepackage{graphicx}
\usepackage{booktabs}
\usepackage{multirow}
\usepackage[table]{xcolor}

\usepackage[accsupp]{axessibility}  

\usepackage{hyperref}

\usepackage{orcidlink}

\begin{document}

\title{MotionAnymesh: Physics-Grounded Articulation for Simulation-Ready Digital Twins} 

\titlerunning{Abbreviated paper title}

\author{WenBo Xu\inst{1} \and
Liu Liu\inst{1} \and
Li Zhang\inst{2} \and
Dan Guo\inst{1} \and
RuoNan Liu\inst{3}}


\institute{Hefei University of Technology \\
The Hong Kong Polytechnic University\\
Shanghai Jiao Tong University}

\maketitle
\begin{figure}[htbp]
    \centering
        \vspace{-1.0em}
    \includegraphics[width=\textwidth]{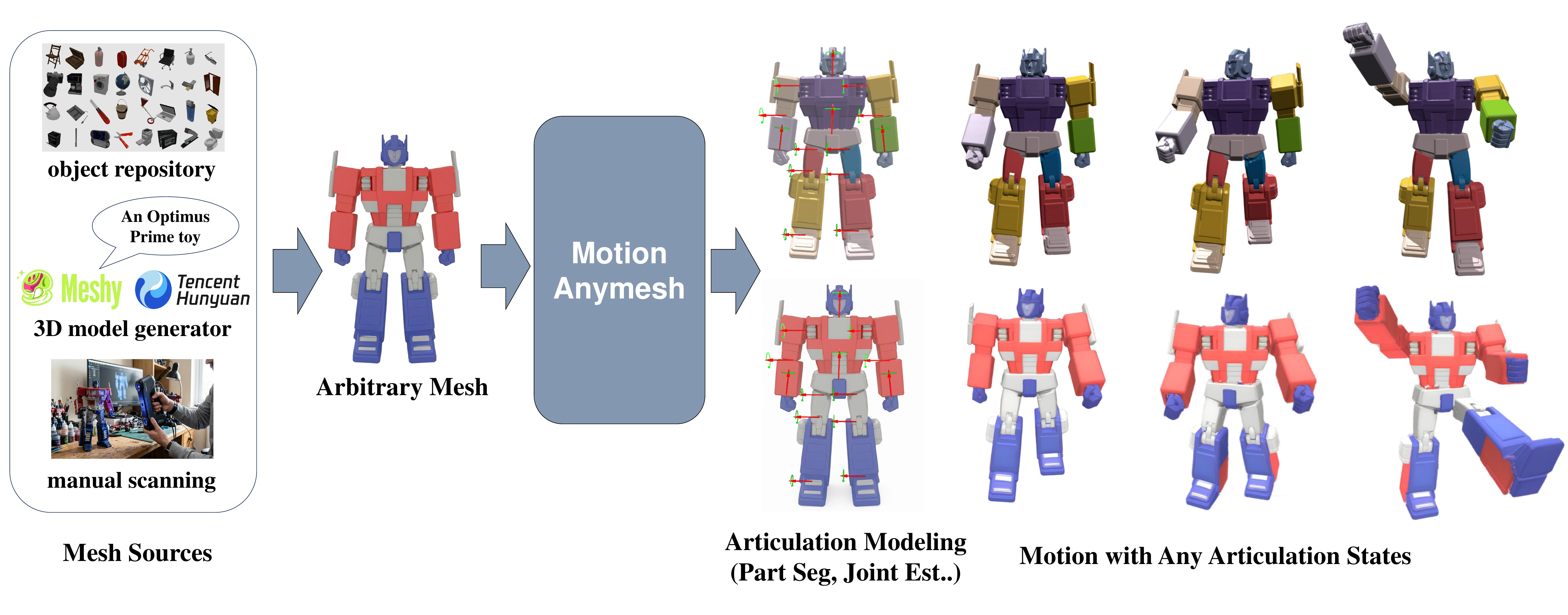}
    \caption{\textbf{MotionAnymesh: Physics-Grounded Articulation of Static 3D Assets.} While current SOTAs (e.g., Articulate-AnyMesh) rely on ungrounded semantics and often suffer from severe inter-penetration, MotionAnymesh ensures physically plausible articulation. Through \textit{kinematic-aware perception} and \textit{physics-constrained optimization}, our zero-shot framework transforms static meshes into collision-free, simulation-ready URDF digital twins for direct deployment in embodied AI tasks.}
    \label{fig:teaser}
    \vspace{-2.5em}
\end{figure}

\vspace{-1.0em}
\begin{abstract}
Converting static 3D meshes into interactable articulated assets is crucial for embodied AI and robotic simulation. However, existing zero-shot pipelines struggle with complex assets due to a critical lack of physical grounding. Specifically, ungrounded Vision-Language Models (VLMs) frequently suffer from kinematic hallucinations, while unconstrained joint estimation inevitably leads to catastrophic mesh inter-penetration during physical simulation. To bridge this gap, we propose \textbf{MotionAnymesh}, an automated zero-shot framework that seamlessly transforms unstructured static meshes into simulation-ready digital twins. Our method features a kinematic-aware part segmentation module that grounds VLM reasoning with explicit SP4D physical priors, effectively eradicating kinematic hallucinations. Furthermore, we introduce a geometry-physics joint estimation pipeline that combines robust type-aware initialization with physics-constrained trajectory optimization to rigorously guarantee collision-free articulation. Extensive experiments demonstrate that MotionAnymesh significantly outperforms state-of-the-art baselines in both geometric precision and dynamic physical executability, providing highly reliable assets for downstream applications.
  \keywords{Articulated Object Modeling \and Kinematic Segmentation \and 3D Assets}
\end{abstract}

\section{Introduction}
\vspace{-0.5em}
\label{sec:intro}
Articulated objects are ubiquitous in human daily environments and interactions. Constructing high-fidelity digital twins of these objects is crucial for applications such as embodied AI \cite{brohan2022rt, duan2022survey, deng2024articulate, deng2025anymate, huang2025mv, kim2025parahome, jiang2024autonomous}, robotic manipulation simulation \cite{xiang2020sapien, gu2023maniskill2, li2021igibson}, and virtual/augmented reality (VR/AR) \cite{kim2025meta}. However, a significant gap exists between the richness of real-world interaction data and the scarcity of assets in simulation environments. In current large-scale open-source 3D asset libraries (such as Objaverse \cite{deitke2023objaverse}), the vast majority of models are purely static meshes, lacking the kinematic structures necessary for interaction, part-level physical boundary segmentation, and precise joint parameters. Traditionally, converting this vast amount of static assets into URDF models that support physically simulated manipulation requires extremely time-consuming and costly manual modeling. Therefore, how to automatically achieve the "static-to-articulated" transformation of 3D assets has become a critical challenge that urgently needs to be addressed in the fields of computer vision and graphics.



Existing pipelines \cite{le2024articulate, qiu2025articulate, lu2025dreamart} for articulated object parsing are predominantly bottlenecked by two fundamental flaws. First, they commonly rely on 2D-to-3D mask lifting strategies (e.g., performing 2D segmentation on rendered views and projecting back to 3D). This view-dependent mechanism inherently fragments the geometric continuity of the 3D shape, inevitably yielding jagged part boundaries and failing completely on self-occluded internal structures. Second, when directly employing Vision-Language Models (VLMs) for open-vocabulary part decomposition, the models rely heavily on semantic priors rather than physical constraints. When faced with complex or irregular mechanical components lacking explicit semantic names, the VLM frequently suffers from severe kinematic hallucinations, either erroneously merging distinct active parts or over-segmenting monolithic structures. Furthermore, existing generative or regression-based approaches \cite{le2024articulate, qiu2025articulate, chen2024urdformer, liu2024singapo, geng2023gapartnet, heppert2023carto, li2020category, wang2019shape2motion} for joint parameter estimation typically lack strict spatial and physical constraints. Although the predicted joint axes or origins may appear plausible in isolation, even a marginal geometric deviation will violently accumulate during long-range actuation. Consequently, when imported into rigorous physics engines like SAPIEN \cite{xiang2020sapien}, these assets frequently violate non-interpenetration constraints, resulting in catastrophic non-physical behaviors such as severe mesh collision, structural detachment, or kinematic freezing. These limitations highlight a critical gap between current visual perception pipelines and genuinely physically executable digital twins.

To address these fundamental limitations, we propose \textbf{MotionAnymesh}, a unified zero-shot framework that reconstructs simulation-ready articulated structures from static meshes (as illustrated in Fig. \ref{fig:teaser}). To eliminate geometric fragmentation while avoiding VLM hallucinations, we strategically decouple boundary extraction from semantic reasoning. We first extract fine-grained geometric primitives directly in the 3D-native space, rigorously preserving sharp and physically valid boundaries. Subsequently, rather than relying on pure semantics, we introduce explicit multi-view kinematic priors derived from SP4D \cite{zhang2025stable}. By grounding the VLM with these physical cues and multi-view observations, we guide it to assemble the 3D-native primitives like following a physical "assembly manual", effectively eradicating kinematic hallucinations. Building upon the recovered part hierarchy, we introduce a rigorous two-stage joint estimation pipeline to guarantee physical executability. Initially, a \textit{Type-Aware Kinematic Initialization} deduces robust initial parameters from localized contact interfaces via PCA and robust 2D RANSAC fitting. Subsequently, a \textit{Physics-Constrained Trajectory Optimization} refines these axes by minimizing unified surface distances. During virtual articulation, any geometric inter-penetration induces strong asymmetric penalties, forcing the joint parameters to systematically correct micro-misalignments and converge toward absolutely valid motion manifolds.

We conduct extensive experiments across diverse benchmarks. Quantitative and qualitative results demonstrate that MotionAnymesh significantly outperforms existing zero-shot baselines, producing geometrically precise and strictly collision-free URDF assets ready for embodied AI downstream tasks. 
The contributions of this work can be summarized as follows:
\vspace{-0.5em}
\begin{enumerate}
    \item We introduce \textbf{MotionAnymesh}, a novel zero-shot pipeline that directly converts static 3D assets into simulation-ready articulated objects, bridging the gap between geometric perception and interactive physical simulation.
    
    \item We propose a robust perception-to-actuation methodology: a Kinematic-Aware Part Segmentation that integrates SP4D priors with VLM reasoning to eliminate hallucinations, and a Geometry-Physics Joint Estimation pipeline, combining robust type-aware initialization with physics-constrained optimization to strictly guarantee collision-free kinematics.
    
    \item We establish a comprehensive evaluation protocol, demonstrating through extensive experiments that our method achieves superior performance.
\end{enumerate}


\vspace{-1.5em}
\section{Related Work}
\vspace{-1.0em}
\subsection{Articulated object modeling}
Existing methods for articulated object modeling broadly fall into three categories. First, visual reconstruction methods \cite{liu2023paris, jiang2022ditto, wu2025reartgs, shen2025gaussianart, guo2025articulatedgs, liu2025building} precisely capture kinematics but require dense spatio-temporal observations (e.g., multi-view or multi-state inputs), limiting their scalability to massive single-state static 3D assets. Second, generative and procedural approaches \cite{liu2024singapo, su2025artformer, chen2024urdformer, liu2024cage, mandi2024real2code} rely on predefined mesh libraries or templates, which hinders their generalization to diverse, open-world shapes. Recently, foundation-model-driven pipelines have enabled open-vocabulary articulation but struggle with complex assets. Methods heavily reliant on Vision-Language Models (VLMs) or 3D LLMs \cite{le2024articulate, qiu2025articulate, yang2025artiworld, li2025urdf} lack physical grounding, making them highly susceptible to kinematic hallucinations when parsing irregular mechanical components without explicit semantic names. Alternatively, approaches leveraging synthesized generative priors (e.g., diffusion or motion videos) \cite{lu2025dreamart, chen2025freeart3d, he2025spark} are prone to geometric inter-penetration and motion distortion under strict 3D topological constraints. Consequently, they fail to provide the high-precision physical guidance required for robust parameter optimization on objects with deeply nested internal structures.
\vspace{-1.0em}
\subsection{Part aware 3D generation}
\vspace{-0.5em}
Instead of generating an 3D object as a whole, part aware 3D generation methods generate objects with part-level information. Recent part-aware 3D generation methods aim to synthesize objects with explicit compositional structures. Early approaches \cite{gao2019sdm, yang2022dsg, mo2019structurenet,wu2020pq} predominantly rely on Variational Autoencoders (VAEs) or graph-based networks to explicitly disentangle global structural topology from detailed part geometries. More recently, diffusion-based paradigms \cite{nakayama2023difffacto, koo2023salad} have gained traction, leveraging powerful generative priors to synthesize high-fidelity part latents. Additionally, recent zero-shot 3D segmentation networks \cite{ma2025p3, yang2024sampart3d, liu2025partfield, liu2023partslip, zhu2025partsam, tang2024segment} excel at identifying geometric boundaries. However, lacking kinematic priors, their operations based purely on local features inevitably lead to the over-segmentation of meshes. To address this lack of kinematic awareness, another line of research specifically targets the kinematic segmentation of articulated objects, recent foundation-model-driven pipelines \cite{qiu2025articulate, le2024articulate, lu2025dreamart} employ Vision-Language Models (VLMs)  to jointly infer semantic parts and kinematic structures. Yet, these paradigms introduce their own bottlenecks. Pipelines relying on 2D-to-3D segmentation lifting inevitably destroy geometric purity, resulting in jagged physical boundaries and failing on heavily occluded interiors. Furthermore, purely semantic-driven segmentation frequently suffers from severe kinematic hallucinations when confronted with fine-grained, irregular mechanical components that lack explicit semantic labels.

\begin{figure}[t]
    \centering
    \includegraphics[width=\textwidth]{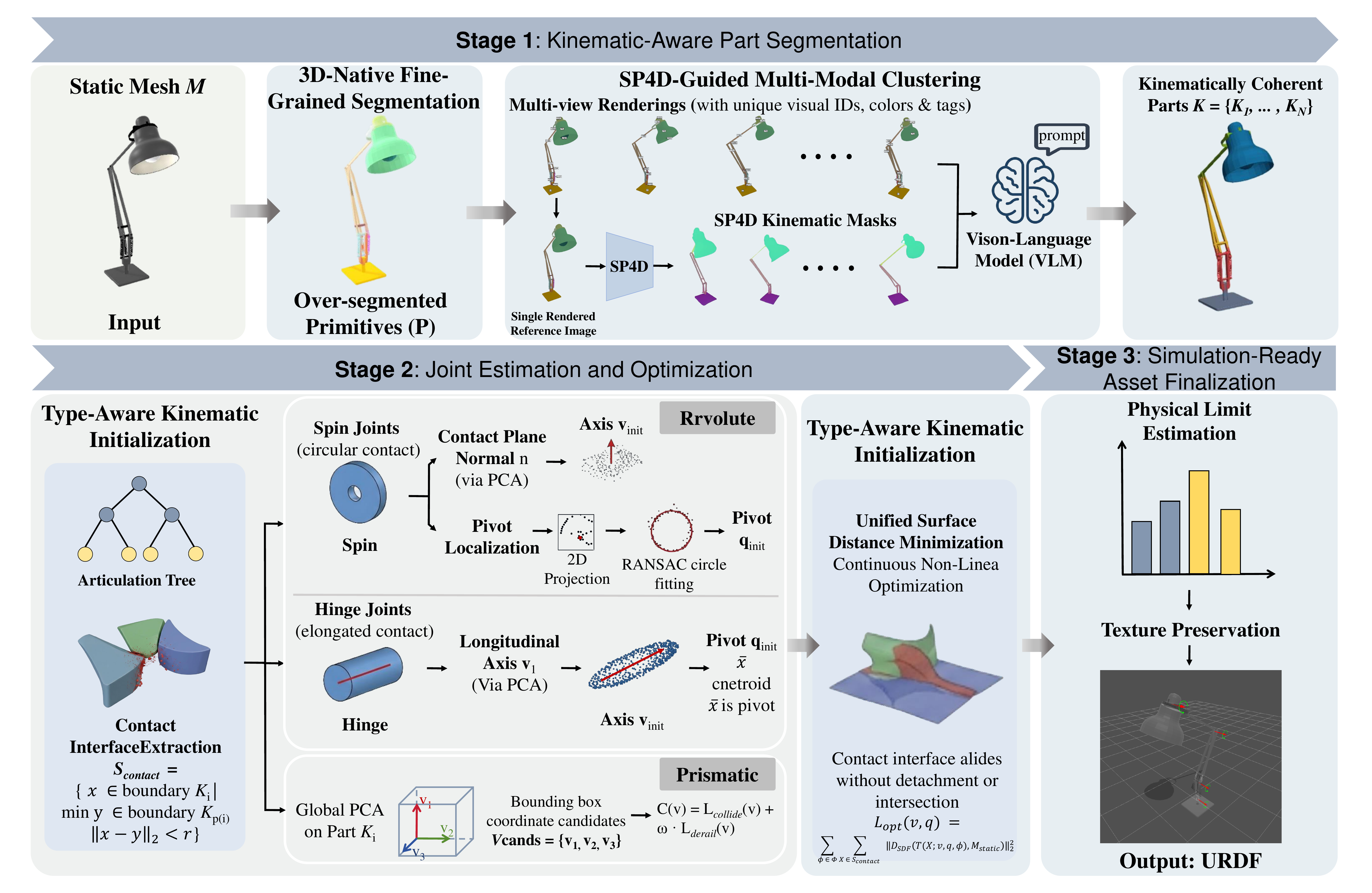}
    \caption{Overview of the \textbf{MotionAnymesh} framework. Our pipeline consists of three integrated stages: (1) Kinematic-Aware Part Segmentation, which extracts 3D-native primitives and clusters them using SP4D kinematic priors and VLM reasoning; (2) Joint Estimation and Optimization, featuring type-aware geometric initialization and physics-constrained trajectory refinement to ensure collision-free articulation ; and (3) Simulation-Ready Asset Finalization, providing simulation-ready URDF models with preserved textures.}
    \vspace{-2.5em}
    \label{fig:pipeline}
\end{figure}

\vspace{-1.0em}
\section{Method}
\vspace{-0.5em}
\subsection{Overview}
\vspace{-0.5em}
As illustrated in Fig. \ref{fig:pipeline}, MotionAnymesh transforms static 3D meshes into simulation-ready articulated digital twins through three integrated stages: (1) \textit{Kinematic-Aware Part Segmentation} (Sec. 3.2), which extracts geometrically pure primitives and clusters them using SP4D-guided VLM reasoning; (2) \textit{Joint Estimation and Optimization} (Sec. 3.3), which deduces robust joint parameters and strictly guarantees collision-free kinematics via unified trajectory optimization; and (3) \textit{Simulation-Ready Asset Finalization} (Sec. 3.4), which determines physical limits and preserves textures to output high-fidelity URDF models.

\vspace{-1.0em}
\subsection{Kinematic-Aware Part Segmentation}
\vspace{-0.5em}
\label{keinmatic seg}
\textbf{\textit{3D-Native Fine-Grained Segmentation.}} To preserve geometric purity, we operate natively in the 3D space. Given the input static mesh $\mathcal{M}$, we employ a 3D-native foundation model, P3-SAM \cite{ma2025p3}, to extract low-level geometric boundaries based on spatial concavities and structural connectivity. This process over-segments the mesh into a set of geometrically pure, disjoint primitives $\mathcal{P} = \{p_1, p_2, \dots, p_m\}$. While these primitives preserve perfect physical boundaries, they lack high-level kinematic semantics (e.g., a single drawer might be fractured into a handle, a front panel, and sliding rails). 

\noindent\textbf{\textit{SP4D-Guided Multi-Modal Clustering.}} To assemble the fragmented primitives $\mathcal{P}$ into kinematically coherent movable parts $\mathcal{K} = \{K_1, \dots, K_N\}$, we formulate a multi-modal clustering process. While large Vision-Language Models (VLMs) demonstrate strong semantic reasoning capabilities, relying exclusively on them often leads to severe kinematic hallucinations, especially when encountering complex mechanical components devoid of explicit semantic labels. Therefore, we introduce explicit kinematic priors derived from SP4D \cite{zhang2025stable} to ground the VLM's reasoning in physical reality.

Specifically, we render the over-segmented mesh from multiple viewpoints, projecting the 3D primitives $\mathcal{P}$ into 2D images and assigning a distinct visual ID (unique colors and numerical tags) to each primitive. Simultaneously, we leverage SP4D to extract explicit kinematic priors. By simply feeding a single rendered reference image of the object into SP4D, it infers and synthesizes consistent multi-view kinematic segmentation masks. These masks explicitly highlight the coarse functional regions of movable parts across different viewing angles. By feeding both the multi-view primitive images (with visual IDs) and the corresponding SP4D kinematic masks into the VLM, we prompt the model to visually correlate the fine-grained geometric IDs with the coarse kinematic regions. We leverage the VLM as a spatial reasoning agent. It cross-references the purely geometric primitives against the explicit physical priors, effectively grouping the fractured pieces into coherent, functional kinematic sets (detailed prompts are provided in the Supplementary Material). The final aggregated parts $K_i = \bigcup_{j \in \mathcal{I}_i} p_j$ (where $\mathcal{I}_i$ is the index set of primitives belonging to the $i$-th kinematic part) possess both pristine 3D geometric boundaries and accurate kinematic hierarchies.

\vspace{-1.0em}
\subsection{Joint Estimation and Optimization}
\vspace{-0.5em}
\label{joint estimation}
With the mesh successfully decomposed into kinematically coherent parts $\mathcal{K} = \{K_0, K_1, \dots, K_N\}$, these isolated components still lack the structural connectivity and motion semantics required for physical simulation. To achieve this, we harness the common-sense reasoning capabilities of the Vision-Language Model (VLM) to construct a comprehensive \textit{articulation tree}. Building upon the visual identifiers assigned during the clustering stage, we prompt the VLM to infer the parent-child dependencies among the parts and to predict the corresponding semantic joint type for each movable component. Based on visual cues and mechanical priors, the VLM broadly classifies the joints into two primary categories: Revolute (further sub-categorized into Spin and Hinge) and Prismatic. Guided by this VLM-derived articulation tree, we deduce the precise physical joint parameters. To ensure physically plausible articulation, we propose a two-stage, type-aware geometric approach: \textit{type-aware kinematic initialization} followed by \textit{physics-constrained trajectory optimization}.

\noindent\textbf{Contact Interface Extraction.} 
Regardless of the joint type, the physical mechanism is intrinsically anchored at the spatial intersection between an articulated part $K_i$ and its parent $K_{P(i)}$. We first extract the contact point cloud $S_{contact}$ by identifying vertices on $K_i$ that are in close proximity to $K_{P(i)}$:
\begin{equation}
S_{contact} = \left\{ \mathbf{x} \in \partial K_i \;\middle|\; \min_{\mathbf{y} \in \partial K_{P(i)}} \|\mathbf{x} - \mathbf{y}\|_2 < \tau \right\}
\end{equation}
where $\partial K$ denotes the surface vertices, and $\tau$ is a predefined distance threshold, empirically set to $0.01$~m.

\noindent\textbf{Type-Aware Kinematic Initialization.} 
With the contact interface extracted, we employ distinct geometric strategies to deduce the initial joint parameters tailored to their specific mechanical behaviors:

\begin{itemize}
    \item \textbf{Revolute Joints:} For rotational mechanisms, the kinematics are fundamentally defined by a rotational axis $\mathbf{v}_{init}$ and a pivot point $q_{init}$. Depending on the geometric nature of the contact interface, we categorize revolute joints into two subtypes and adopt tailored initialization strategies:
    \begin{itemize}
        \item \textit{Spin Joints:} For spin mechanisms (e.g., wheels, knobs), the visible mechanical connection is typically characterized by an axially symmetric intersection boundary. In most surface meshes, the extracted contact point cloud $S_{contact}$ between the spin part and its base forms a disc-like or shallow annular band. By applying Principal Component Analysis (PCA) to $S_{contact}$, we evaluate its spatial variance. Since these localized intersection points distribute predominantly along the radial cross-section rather than the axial depth, the eigenvector corresponding to the smallest eigenvalue $\mathbf{n}$ robustly captures the orthogonal normal of this distribution. We assign this normal directly as the initial rotational axis ($\mathbf{v}_{init} = \mathbf{n}$).
        
        To accurately localize the rotational pivot $q_{init}$ against surface noise, we propose a robust 2D-projection and RANSAC-based fitting pipeline. Let $\bar{\mathbf{x}}$ be the geometric centroid of $S_{contact}$. We establish a local 2D orthogonal coordinate system using two unit vectors $\mathbf{b}_1$ and $\mathbf{b}_2$, both perpendicular to $\mathbf{n}$. Every 3D contact point $\mathbf{x} \in S_{contact}$ is projected into this 2D local space to obtain its 2D coordinates $\mathbf{p}$:
        \begin{equation}
        \mathbf{p} = \begin{bmatrix} (\mathbf{x} - \bar{\mathbf{x}})^T \mathbf{b}_1 \\ (\mathbf{x} - \bar{\mathbf{x}})^T \mathbf{b}_2 \end{bmatrix}
        \end{equation}
        
        Geometrically, the contact boundary of a spin joint must exhibit rotational symmetry around its axis. Consequently, when projected onto the local 2D plane perpendicular to $\mathbf{n}$, the valid contact points naturally distribute along a circular locus. To rigorously deduce this center while rejecting outliers and handling partial observations, we employ the RANSAC algorithm to iteratively fit a 2D circle to the projected points $\mathbf{p} = \{\mathbf{p}_i\}_{i=1}^N$. We evaluate the hypothesized circle (center $\mathbf{c}_{2D}$, radius $r$) by maximizing the consensus set (inliers):
        
        \begin{equation}
        \mathbf{c}_{2D}^*, r^* = \arg\max_{\mathbf{c}_{2D}, r} \sum_{i=1}^{N} \mathbb{I} \left( \Big| \|\mathbf{p}_i - \mathbf{c}_{2D}\|_2 - r \Big| < \delta \right)
        \end{equation}
        
        where $\mathbb{I}(\cdot)$ is the indicator function and $\delta$ is the inlier distance threshold that accommodates surface noise and quantization errors. In our implementation, $\delta$ is empirically set to $0.005$~m. Finally, the optimal 2D center $\mathbf{c}_{2D}^* = [x_c, y_c]^T$ is mapped back into the global 3D space to serve as the highly accurate joint pivot $q_{init}$:
        
        \begin{equation}
        q_{init} = \bar{\mathbf{x}} + x_c \mathbf{b}_1 + y_c \mathbf{b}_2
        \end{equation}
        
    
    \item \textit{Hinge Joints:} Unlike spin mechanisms, the contact regions of hinge joints (e.g., door hinges, laptop screens) are characteristically distributed along the axis of rotation, exhibiting a dominant longitudinal span. This holds true whether the physical connection is a continuous strip or comprises multiple discrete anchor points. Based on this mechanical prior, we propose a robust PCA-based initialization. We compute the covariance matrix of the zero-centered contact point cloud $S_{contact}$. Regardless of the local patch shapes, the primary eigenvector $\mathbf{v}_1$ (corresponding to the largest eigenvalue) intrinsically captures the direction of maximum spatial variance of the entire point set, perfectly representing the global longitudinal hinge line. Thus, we assign the initial rotational axis as $\mathbf{v}_{init} = \mathbf{v}_1$. Furthermore, because such contact distributions are generally symmetric along their rotational axis, we directly assign the geometric centroid $\bar{\mathbf{x}}$ of the point cloud as the optimal mechanical pivot ($q_{init} = \bar{\mathbf{x}}$).
    \end{itemize}

\item \textbf{Prismatic Joints:} Unlike revolute joints, the sliding axis of a prismatic joint is dictated by the global geometry of the part rather than the localized contact patch. We first perform global PCA on the entire point cloud of $K_i$ to boldly extract its three local bounding-box coordinate axes $\mathcal{V}_{cands} = \{\mathbf{v}_1, \mathbf{v}_2, \mathbf{v}_3\}$ as candidates. 
    
To determine the true sliding direction among these candidates, we propose a \textit{Normalized Dual-Penalty Trajectory Verification} mechanism. The physical intuition is that a correctly estimated prismatic axis must allow the part to slide without penetrating the parent body (collision-free) while strictly adhering to its mechanical rails (derailment-free). We define a set of discrete translation steps $\mathcal{D} = \{d_1, d_2, \dots, d_M\}$ covering both forward and backward movements. For each candidate axis $\mathbf{v} \in \mathcal{V}_{cands}$ and step $d \in \mathcal{D}$, the virtually translated part is denoted as $K_i(d, \mathbf{v}) = \{\mathbf{x} + d\mathbf{v} \mid \mathbf{x} \in K_i\}$.
    
We evaluate a comprehensive scoring function $\mathcal{C}(\mathbf{v})$ that penalizes both geometric collision ($\mathcal{L}_{collide}$) and mechanical derailment ($\mathcal{L}_{derail}$):
    
\begin{equation}
\mathcal{C}(\mathbf{v}) = \mathcal{L}_{collide}(\mathbf{v}) + \omega \cdot \mathcal{L}_{derail}(\mathbf{v})
\end{equation}
    
Specifically, the collision penalty $\mathcal{L}_{collide}$ computes the normalized ratio of points on the moved part that geometrically inter-penetrate the static parent $K_{P(i)}$. It is defined via an indicator function $\mathbb{I}(\cdot)$ to count penetration events:

\begin{equation}
\mathcal{L}_{collide}(\mathbf{v}) = \frac{1}{|\mathcal{D}| |K_i|} \sum_{d \in \mathcal{D}} \sum_{\mathbf{x}' \in K_i(d, \mathbf{v})} \mathbb{I} \left( \min_{\mathbf{y} \in K_{P(i)}} \| \mathbf{x}' - \mathbf{y} \|_2 < \epsilon_c \right)
\end{equation}

where $\epsilon_c$ is a strict distance threshold to robustly define a penetration event. In our implementation, $\epsilon_c$ is empirically set to $0.005$~m. Conversely, the derailment penalty $\mathcal{L}_{derail}$ ensures the moving part does not detach from its original sliding track. It measures the average absolute divergence distance from the original stationary contact points $\mathcal{S}_{contact}$ to the newly translated surface:

\begin{equation}
\mathcal{L}_{derail}(\mathbf{v}) = \frac{1}{|\mathcal{D}| |\mathcal{S}_{contact}|} \sum_{d \in \mathcal{D}} \sum_{\mathbf{p} \in \mathcal{S}_{contact}} \min_{\mathbf{x}' \in K_i(d, \mathbf{v})} \| \mathbf{p} - \mathbf{x}' \|_2
\end{equation}

To equalize the magnitude differences between the dimensionless probability $\mathcal{L}_{collide}$ and the absolute spatial deviation $\mathcal{L}_{derail}$, $\omega$ serves as a crucial balancing hyperparameter. We empirically set $\omega = 20$ to align their scales. The axis $\mathbf{v}^* = \arg\min_{\mathbf{v} \in \mathcal{V}_{cands}} \mathcal{C}(\mathbf{v})$ that yields the minimum dual-penalty cost is selected as the definitive sliding direction. Pivot points are mathematically undefined and unnecessary for prismatic joints.
\end{itemize}

\noindent\textbf{Physics-Constrained Trajectory Optimization.}
Although the initialized parameters provide a plausible mechanical layout, subtle surface misalignments (e.g., PCA-derived axes slightly tilted relative to true rails) may induce friction or penetration during long-range motion. To guarantee simulation-ready assets, we formulate a continuous non-linear optimization to refine joint parameters.

To achieve this, we enforce a \textit{Unified Surface Distance Minimization} constraint across all joint types. The core physical intuition is that during articulation, the contact interface $S_{contact}$ should act as a perfect mechanical bearing or sliding rail, seamlessly gliding against the parent surface without detachment or intersection. Let $\mathcal{T}(\mathbf{x}; \mathbf{v}, \mathbf{q}, \phi)$ denote the 3D rigid transformation applied to a contact point $\mathbf{x} \in S_{contact}$ during motion. For Revolute joints, $\phi = \theta$ represents a rotation angle around $\mathbf{v}$ anchored at $\mathbf{q}$. For Prismatic joints, $\phi = d$ represents a translation distance along $\mathbf{v}$, reducing the transformation to $\mathcal{T}(\mathbf{x}; \mathbf{v}, d) = \mathbf{x} + d \cdot \mathbf{v}$.

To prevent geometric inter-penetration and maintain tight kinematic constraints, we simulate the part's motion across a set of discrete virtual states $\Phi$ (representing angular steps $\Theta$ for revolute joints, or spatial steps $D$ for prismatic joints). We minimize the unified trajectory deviation loss $\mathcal{L}_{opt}$:

\begin{equation}
\mathcal{L}_{opt}(\mathbf{v}, \mathbf{q}) = \sum_{\phi \in \Phi} \sum_{\mathbf{x} \in S_{contact}} \left\| \mathcal{D}_{SDF} \big( \mathcal{T}(\mathbf{x}; \mathbf{v}, \mathbf{q}, \phi), \mathcal{M}_{static} \big) \right\|_2^2
\end{equation}

where $\mathcal{D}_{SDF}(\cdot, \cdot)$ computes the Signed Distance Field (SDF) relative to the static environment $\mathcal{M}_{static}$. Optimized via the Levenberg-Marquardt algorithm, this unified term rigorously ensures that the moving part maintains minimal, uniform proximity to the static base throughout its entire range of motion, guaranteeing physically valid, collision-free kinematics for both rotational and translational mechanisms.

\vspace{-1.0em}
\subsection{Simulation-Ready Asset Finalization}
\label{post-processing}

To render the generated digital twin fully simulation-ready, two critical final steps are required: determining the valid Range of Motion (RoM) for each joint and finalizing the asset's visual fidelity for standard URDF assembly.

\textbf{Physical Limit Estimation.} 
Unconstrained joints can lead to unrealistic behaviors in physics simulators, such as doors rotating into walls or drawers flying out of cabinets. We employ a forward-simulation collision detection mechanism to automatically deduce the kinematic limits. 
For \textit{Revolute joints}, starting from the rest state ($0^\circ$), we incrementally rotate the movable part in both positive and negative directions up to a maximum of $\pm 180^\circ$. The joint limits $[\theta_{min}, \theta_{max}]$ are strictly defined by the exact angles where a geometric collision (evaluated via mesh intersection) occurs with the static base. 
For \textit{Prismatic joints}, the physical constraints are two-fold. In the inward pushing direction, the limit is typically bounded by collision (e.g., a drawer hitting the cabinet's backplate). In the outward pulling direction, we propose a novel \textit{contact-loss} criterion. We incrementally translate the part along its optimized axis $\mathbf{v}^*$ and continuously monitor the mutual contact area with the parent part. The maximum extension limit $d_{max}$ is determined at the point where this contact area drops to zero, physically simulating the moment a sliding component detaches from its rails.

\noindent\textbf{Texture Preservation and Generative Re-texturing.}
To ensure visual fidelity, our pipeline inherently preserves the original UV mapping and texture information of the input mesh. If the source asset is textured, the decomposed parts naturally retain their original appearance in the final output. However, because true 3D segmentation inevitably exposes texture-less internal cross-sections, and to support customized visual editing, we introduce an optional generative re-texturing module. Leveraging state-of-the-art 3D generative models like Hunyuan3D \cite{zhao2025hunyuan3d}, we can independently generate entirely new, high-fidelity textures for each isolated part based on user-defined text prompts. Ultimately, these textured components, alongside their optimized kinematic parameters, are automatically assembled into a standard URDF format, delivering collision-free, photorealistic articulated assets ready for immediate deployment in interactive simulation environments.

\vspace{-1.5em}
\section{Experiments}
\vspace{-1.0em}
\noindent\textbf{Implementation Details.}
We employ P3-SAM \cite{ma2025p3} for 3D-native geometric primitive extraction and SP4D \cite{zhang2025stable} to generate multi-view kinematic masks, with GPT-4o serving as our core VLM. For trajectory optimization, we utilize Trimesh to compute the SDF and solve the non-linear objective via SciPy's Nelder-Mead algorithm. Hunyuan3D \cite{zhao2025hunyuan3d} is integrated for optional generative re-texturing. Finally, the optimized parameters and processed meshes are compiled into simulation-ready URDF assets. All experiments are executed on three NVIDIA RTX 4090 GPUs.

\noindent\textbf{Datasets.} To evaluate our zero-shot framework, we construct a test suite from three sources: (1) \textbf{PartNet-Mobility} \cite{xiang2020sapien}, providing standard ground-truth URDF annotations for exact metric calculations; (2) \textbf{Objaverse} \cite{deitke2023objaverse}, featuring diverse, open-vocabulary static meshes without predefined templates; and (3) \textbf{Generative 3D Assets} from recent Text/Image-to-3D models to stress-test robustness. To enable quantitative comparisons on the latter two unstructured domains, we manually authored their ground-truth URDFs, including part masks and joint parameters.

\noindent\textbf{Baselines.} We compare against five state-of-the-art 3D articulation approaches: Articulate-Anything \cite{le2024articulate}, Articulate-AnyMesh \cite{qiu2025articulate}, URDFormer \cite{chen2024urdformer}, SINGAPO \cite{liu2024singapo}, and PARIS \cite{liu2023paris}.

\noindent{\textbf{Metrics.}} To evaluate the generated articulated digital twins, we employ a combination of static geometric metrics and dynamic physical simulation tests: \textbf{1) Part Segmentation Quality:} We measure the \textbf{mean Intersection over Union (mIoU)} between the predicted movable parts and ground-truth meshes to assess geometric precision. Additionally, we report \textbf{Count Accuracy (Count Acc)}, defined as the percentage of test instances where the inferred number of active kinematic links perfectly matches the ground truth, which validates the topological reasoning capability of our VLM-guided clustering. \textbf{2) Joint Parameter Estimation:} We benchmark the estimated kinematic parameters against ground-truth URDF annotations using three metrics: (i) \textbf{Joint Type Error}: a binary metric indicating the misclassification of the joint type; (ii) \textbf{Joint Axis Error}: the angular deviation between predicted and ground-truth joint axes, normalized within $[0, \pi]$; and (iii) \textbf{Joint Pivot Error}: the $L_2$ Euclidean distance between predicted and actual joint pivot locations. \textbf{3) Physical Executability:} To guarantee dynamic stability beyond static parameters, we load the generated URDF assets into the SAPIEN physics simulator \cite{xiang2020sapien}. Physical executability is defined as the success rate of URDFs that can be fully actuated along their valid ranges without exhibiting catastrophic non-physical behaviors, such as severe mesh inter-penetration, structural detachment, or kinematic freezing. This critically verifies the asset's suitability for downstream sim-to-real applications.

\vspace{-1.0em}
\subsection{Comparison with Baselines}
\begin{figure}[!htbp]
    \centering
    \vspace{-1.0em}
    \includegraphics[width=\textwidth]{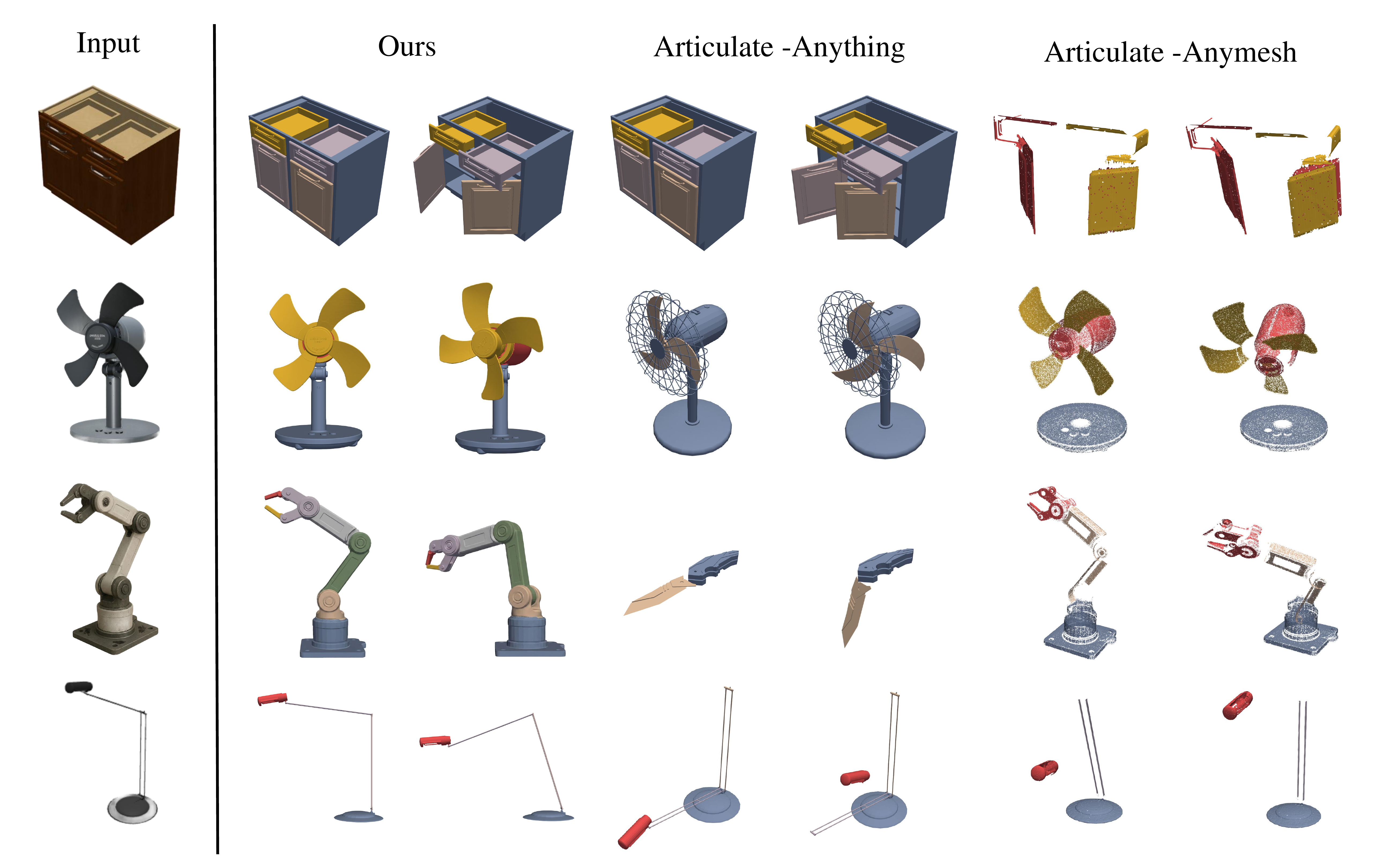}
    \caption{Qualitative comparison of articulated object modeling. Compared with Articulate-Anything and Articulate-AnyMesh, MotionAnymesh produces cleaner geometric boundaries and more accurate kinematic structures, especially for complex mechanical objects like the robot arm and multi-part lamps.}
    \vspace{-2.0em}
    \label{fig:qualitative_comparison}
\end{figure}

\vspace{-0.5em}
We comprehensively evaluate MotionAnymesh against state-of-the-art articulation frameworks, including Articulate-Anything \cite{le2024articulate}, Articulate-AnyMesh \cite{qiu2025articulate}, SINGAPO \cite{liu2024singapo}, URDFormer \cite{chen2024urdformer}, and PARIS \cite{liu2023paris}. The quantitative results across diverse 3D assets are summarized in Table \ref{tab:main_results}. Our qualitative comparisons in Figure \ref{fig:qualitative_comparison} focus on the most competitive zero-shot foundation-model-driven baselines (Articulate-Anything and Articulate-AnyMesh).

\begin{table}[htbp]
\centering
\caption{Quantitative comparison against state-of-the-art baselines across diverse 3D assets. We evaluate our method against others on three core dimensions: Part Segmentation, Joint Parameter Estimation, and Physical Executability. $\uparrow$ indicates higher is better, and $\downarrow$ indicates lower is better. Best results are highlighted in \textbf{bold}.}
\label{tab:main_results}
\resizebox{\textwidth}{!}{
\begin{tabular}{l ccc ccc c}
\toprule
\multirow{2}{*}{\textbf{Method}} & \multicolumn{2}{c}{\textbf{Part Segmentation}} & \multicolumn{3}{c}{\textbf{Joint Estimation}} & \textbf{Simulation} \\
\cmidrule(lr){2-3} \cmidrule(lr){4-6} \cmidrule(lr){7-7}
& mIoU $\uparrow$ & Count Acc $\uparrow$ & Type Err $\downarrow$ & Axis Err $\downarrow$ & Pivot Err $\downarrow$ & Executability $\uparrow$ \\
\midrule
PARIS \cite{liu2023paris} & 0.17 & 0.23 & 0.67 & 1.56 & 1.14 & 11\% \\
URDFormer \cite{chen2024urdformer}             & 0.21 & 0.33 & 0.72 & 1.31 & 1.53 & 21\% \\
SINGAPO \cite{liu2024singapo} & 0.52 & 0.66 & 0.24 & 0.73 & 0.57 & 43\% \\
Articulate-Anything \cite{le2024articulate} & 0.47 & 0.61 & 0.21 & 0.86 & 0.64 & 46\% \\
Articulate-AnyMesh \cite{qiu2025articulate} & 0.59 & 0.74 & 0.35 & 0.64 & 0.44 & 35\% \\
\midrule
\textbf{MotionAnymesh (Ours)}                    & \textbf{0.86} & \textbf{0.92} & \textbf{0.08} & \textbf{0.12} & \textbf{0.10} & \textbf{87\%} \\
\bottomrule
\end{tabular}
}
\vspace{-1.5em}
\end{table}

\noindent\textbf{Part Segmentation.} 
MotionAnymesh significantly outperforms all baselines in geometric decomposition, achieving an mIoU of \textbf{0.86} and a Count Accuracy of \textbf{0.92}, surpassing the strongest baseline (Articulate-AnyMesh) by massive absolute margins of \textbf{+0.27} and \textbf{+0.18}, respectively. Existing retrieval-based methods (URDFormer, Articulate-Anything, SINGAPO) are fundamentally bottlenecked by predefined CAD libraries. As illustrated by Articulate-Anything in Fig. \ref{fig:qualitative_comparison}, while such paradigms succeed on in-domain objects (e.g., the \textit{storage}), they catastrophically fail on novel geometries like the \textit{robot\_arm}, yielding severe boundary mismatches or entirely irrelevant structures. Conversely, the zero-shot Articulate-AnyMesh relies heavily on open-vocabulary VLM reasoning, suffering from severe over-segmentation on irregular internal mechanisms lacking explicit semantic names. In contrast, our 3D-native approach, grounded by explicit SP4D kinematic priors, rigorously ensures topologically sound and geometrically intact segmentation across open-world assets.

\noindent\textbf{Joint Estimation and Physical Executability.}
Regarding kinematic alignment, MotionAnymesh drastically reduces geometric deviations, slashing Axis Error down to \textbf{0.12} and Pivot Error to \textbf{0.10}. Retrieval-based baselines frequently fail at spatial composition even when correct parts are retrieved (e.g., the \textit{lamp} in Fig. \ref{fig:qualitative_comparison}), leading to completely erroneous joint parameters. Meanwhile, Articulate-AnyMesh attempts to estimate joints via 2D-to-3D VLM projection heuristics, making it highly susceptible to 3D spatial hallucinations and severe axis deviations. MotionAnymesh elegantly circumvents these bottlenecks by anchoring kinematics in robust 3D-native geometry. Our Type-Aware Kinematic Initialization deduces reliable parameters directly from localized contact interfaces, while our Physics-Constrained Trajectory Optimization strictly penalizes geometric inter-penetration.  Consequently, our pipeline not only minimizes joint errors but achieves an unprecedented $\textbf{87\%}$ Physical Executability, nearly doubling the success rate of the optimal method (Articulate-Anything at $46\%$).

\vspace{-1.0em}
\subsection{Ablation Study}
To rigorously validate the individual contributions of our proposed core modules, we systematically disable specific components of MotionAnymesh and evaluate the degraded variants. The decoupled ablation results for part segmentation and joint estimation are presented in Table \ref{tab:ablation_seg} and Table \ref{tab:ablation_opt}, respectively.

\begin{table}[htbp]
\centering
\caption{\textbf{Ablation on SP4D kinematic priors.} Removing SP4D guidance ("w/o SP4D") causes the VLM to rely solely on semantic logic, leading to severe kinematic hallucinations and over-segmentation.}
\label{tab:ablation_seg}
\resizebox{0.75\textwidth}{!}{
\begin{tabular}{l | cc}
\toprule
\textbf{Clustering Strategy} & mIoU $\uparrow$ & Count Acc $\uparrow$ \\
\midrule
Pure VLM Semantics (w/o SP4D) & 0.68 & 0.81 \\
\textbf{SP4D-Guided Clustering (Ours)} & \textbf{0.86} & \textbf{0.92} \\
\bottomrule
\end{tabular}
}
\vspace{-1.0em}
\end{table}

\noindent\textbf{Effectiveness of SP4D Kinematic Priors.} 
We introduce a "w/o SP4D" variant by removing multi-view kinematic masks during part clustering, forcing the VLM to aggregate 3D-native primitives based purely on visual semantics. As shown in Table \ref{tab:ablation_seg}, this purely semantic approach leads to a drastic drop in Count Accuracy and mIoU. Semantic understanding alone is insufficient for articulation modeling; without explicit kinematic priors, the VLM frequently suffers from severe kinematic hallucinations, either erroneously merging distinct but visually similar components, or over-segmenting monolithic irregular structures lacking explicit semantic labels. In contrast, our SP4D-guided clustering strictly bounds the VLM's reasoning within physical reality, ensuring geometrically intact and topologically accurate part extraction.

\begin{table}[htbp]
\centering
\vspace{-1.0em}
\caption{\textbf{Impact of physics-constrained optimization.} While geometric initialization provides a reasonable starting point, our non-linear optimization is essential to eliminate micro-misalignments and guarantee high physical executability in simulators.}
\label{tab:ablation_opt}
\resizebox{0.95\textwidth}{!}{
\begin{tabular}{l | cc | c}
\toprule
\textbf{Joint Estimation Method} & Axis Err $\downarrow$ & Pivot Err $\downarrow$ & Executability $\uparrow$ \\
\midrule
Kinematic Initialization (w/o Opt.) & 0.23 & 0.27 & 65\%\\
\textbf{Physics-Constrained Opt. (Ours)} & \textbf{0.12} & \textbf{0.10} & \textbf{87\%} \\
\bottomrule
\end{tabular}
}
\vspace{-1.5em}
\end{table}

\noindent\textbf{Necessity of Physics-Constrained Optimization.} 
The "w/o Opt." variant bypasses the non-linear trajectory refinement, directly outputting the initialized joint parameters. As shown in Table \ref{tab:ablation_opt}, while achieving seemingly reasonable static metrics, it suffers a catastrophic collapse in dynamic Physical Executability. Purely geometric initialization only provides a macro-level orthogonal guess; even a marginal angular deviation violently accumulates during long-range articulation, inevitably causing severe mesh inter-penetration or structural freezing in physics simulators. By actively penalizing geometric collisions and minimizing surface distances, our optimization systematically corrects these micro-misalignments, guaranteeing collision-free and seamlessly executable digital twins.

\vspace{-1.0em}
\subsection{Applications}

To demonstrate the versatility and physical reliability of MotionAnymesh, we showcase its integration into two downstream workflows: open-vocabulary asset generation and embodied robotic manipulation.

\begin{figure}[htbp]
    \centering
    \includegraphics[width=\textwidth]{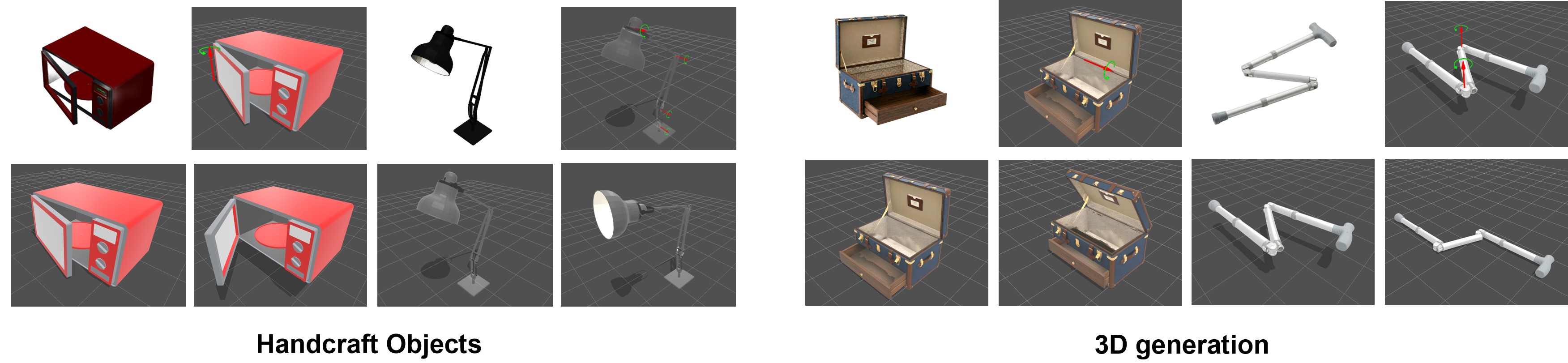}
    \caption{Versatile articulation across diverse domains. MotionAnymesh successfully processes handcrafted assets and AI-generated surface meshes, transforming unstructured geometry into interactive digital twins without manual intervention.}
    \label{fig:app_diverse_assets}
    \vspace{-1.0em}
\end{figure}

\noindent\textbf{Versatile Articulation for Diverse 3D Assets.} 
MotionAnymesh serves as a universal articulation engine across diverse 3D domains (Fig. \ref{fig:app_diverse_assets}). For existing artist-designed datasets (e.g., PartNet-Mobility), it automatically extracts part segmentations and deduces precise joint structures without manual annotations. Furthermore, by processing static surface meshes synthesized by state-of-the-art 3D generative models, MotionAnymesh effortlessly transforms unstructured, open-vocabulary AI assets into high-fidelity, interactable URDF models ready for interactive simulation.

\begin{figure}[htbp]
    \centering
    \includegraphics[width=\textwidth]{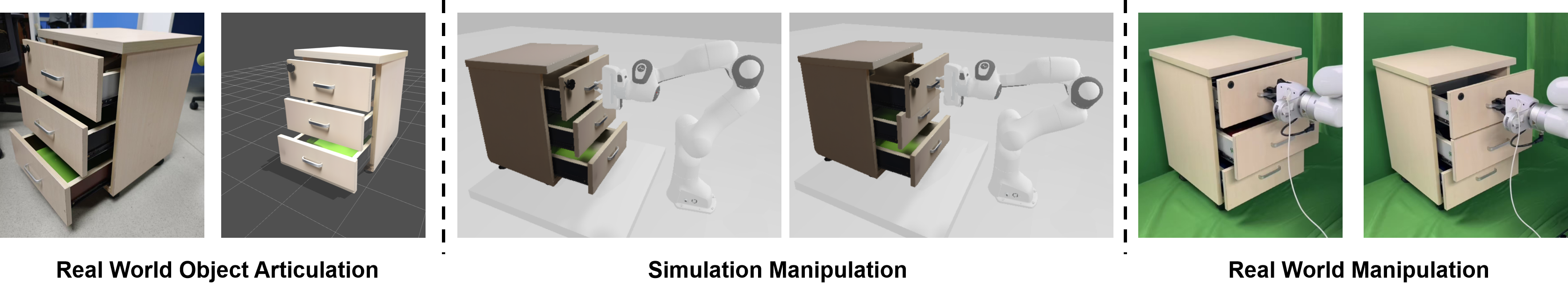}
    \caption{Real-to-Sim-to-Real application. (Left) A static mesh is reconstructed from a single image. (Middle) MotionAnymesh generates a collision-free URDF for policy learning in simulation. (Right) The learned manipulation policy is deployed onto a physical robot, validating the sim-to-real fidelity of our estimated parameters.}
    \vspace{-1.0em}
    \label{fig:app_sim2real}
\end{figure}

\noindent\textbf{Real-to-Sim-to-Real Robotic Manipulation.} 
To validate the physical executability of our digital twins, we deploy MotionAnymesh in a complete Real-to-Sim-to-Real pipeline (Fig. \ref{fig:app_sim2real}). Starting from a single real-world photograph, Hunyuan3D \cite{zhao2025hunyuan3d} reconstructs a static 3D surface mesh. MotionAnymesh then processes this raw geometry into a strictly collision-free, articulated URDF. After a robotic arm learns manipulation policies on this interactive asset in simulation, the learned trajectory is successfully deployed onto a physical robot. This end-to-end execution strongly validates that our physics-constrained pipeline produces highly accurate, sim-to-real-ready joint parameters and segmentations.

\vspace{-1.5em}
\section{Conclusion}
\vspace{-1.0em}
\label{sec:conclusion}
In this paper, we presented \textbf{MotionAnymesh}, an automated zero-shot framework that directly transforms unstructured static 3D meshes into simulation-ready articulated digital twins. Addressing the critical lack of physical grounding in existing perception pipelines, we introduced a robust perception-to-actuation methodology. Our kinematic-aware part segmentation grounds Vision-Language Models with explicit SP4D physical priors, effectively eradicating kinematic hallucinations. Furthermore, our geometry-physics joint estimation pipeline enforces strict physics-constrained trajectory optimization to guarantee collision-free kinematics. Extensive evaluations demonstrate that MotionAnymesh significantly outperforms state-of-the-art baselines in both geometric precision and dynamic physical executability, seamlessly bridging the gap between static geometric perception and interactive physical simulation.

%
%
\bibliographystyle{splncs04}
\bibliography{main}
\end{document}